\documentclass[letterpaper]{article} 
\usepackage{aaai2026}  
\usepackage{times}  
\usepackage{helvet}  
\usepackage{courier}  
\usepackage[hyphens]{url}  
\usepackage{graphicx} 
\usepackage{natbib}  
\usepackage{caption} 
\usepackage{algorithm}
\usepackage{algorithmic}

\usepackage{booktabs}
\usepackage{mathtools}
\usepackage{amsfonts}
\usepackage{amsmath}
\usepackage{amsthm}
\usepackage{multirow}
\usepackage{breqn}
\usepackage{amssymb}

\usepackage{newfloat}
\usepackage{listings}
\DeclareCaptionStyle{ruled}{labelfont=normalfont,labelsep=colon,strut=off} 
\floatstyle{ruled}
\newfloat{listing}{tb}{lst}{}
\floatname{listing}{Listing}
\title{Rethinking Saliency Maps: A Cognitive Human Aligned Taxonomy and Evaluation Framework for Explanations}

\author {
    Yehonatan Elisha\textsuperscript{\rm 1},
    Seffi Cohen\textsuperscript{\rm 2},
    Oren Barkan\textsuperscript{\rm 3},
    Noam Koenigstein\textsuperscript{\rm 1}
}
\affiliations {
    \textsuperscript{\rm 1}Tel Aviv University\\
    \textsuperscript{\rm 2}Harvard University\\
    \textsuperscript{\rm 3}The Open University\\
}

\usepackage{bibentry}

\begin{document}

\maketitle

\begin{abstract}
Saliency maps have become a cornerstone of visual explanation in deep learning, yet there remains no consensus on their intended purpose and their alignment with specific user queries. This fundamental ambiguity undermines both the evaluation and practical utility of explanation methods. In this paper, we introduce the Reference-Frame$
\times$Granularity (RFxG) taxonomy—a principled framework that addresses this ambiguity by conceptualizing saliency explanations along two essential axes: the \textit{reference-frame} axis (distinguishing between pointwise "Why Husky?" and contrastive "Why Husky and not Shih-tzu?" explanations) and the \textit{granularity} axis (ranging from fine-grained class-level to coarse-grained group-level interpretations, e.g., “Why Husky?” vs. “Why Dog?”). Through this lens, we identify critical limitations in existing evaluation metrics, which predominantly focus on pointwise faithfulness while neglecting contrastive reasoning and semantic granularity. To address these gaps, we propose four novel faithfulness metrics that systematically assess explanation quality across both RFxG dimensions. Our comprehensive evaluation framework spans ten state-of-the-art methods, 4 model architectures, and 3 datasets. By suggesting a shift from model-centric to user-intent-driven evaluation, our work provides both the conceptual foundation and practical tools necessary for developing explanations that are not only faithful to model behavior but also meaningfully aligned with human understanding.

\end{abstract}
\vspace{-2mm}
\begin{links}
    \link{Code}{https://github.com/yonisGit/RFxG}
\end{links}

\section{Introduction}
\label{sec:intro}

\begin{figure}[t]
\centering
\includegraphics[width=0.5\textwidth]{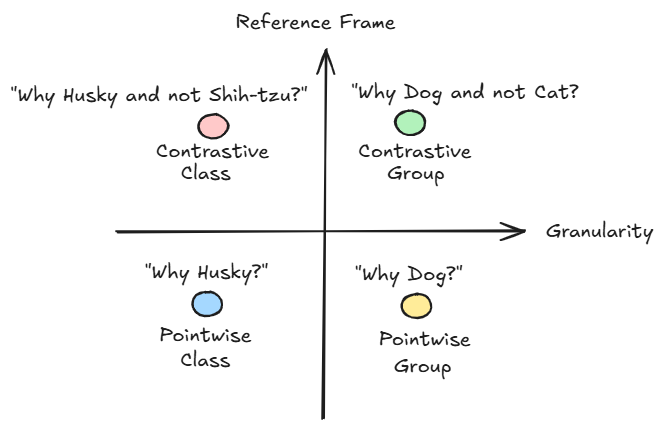}
\vspace{-4mm}
\caption{Our RFxG Explanation Axes. Note that there are also explanations between the points. For example: class-group contrastive questions like "Why Husky and not other Dogs?"}
\label{fig:xai_axis}
\vspace{-3mm}
\end{figure}

The increasing adoption of deep learning systems in high-stakes applications has amplified the need for transparency and interpretability. As deep models grow in complexity, so does the demand for methods that can explain their predictions in ways that are intelligible and actionable to human users. This growing field, known as Explainable AI (XAI), aims to bridge the gap between black-box models and human understanding.

Across domains, numerous works have advanced the study of XAI~\cite{sundararajan2017axiomatic, smilkov2017smoothgrad,chefer2021transformer,barkan2020explainable,barkan2021grad,barkan2024improving, barkan2024learning, barkan2024llm, barkan2025bee}. In computer vision, XAI efforts have primarily focused on visual explanation techniques, with saliency maps being one of the most dominant paradigms~\cite{selvaraju2017grad, chefer2021transformer,barkan2021gam,barkan2023stochastic,barkan2023learning,haddad2025soft}. 
In parallel, contrastive explanation techniques~\cite{dhurandhar2018explanations,xie2023two,wang2023counterfactual} have emerged, aiming to answer comparative questions such as “Why class A rather than class B?”, rather than simply “Why class A?”.

Despite their popularity, the interpretability and trustworthiness of saliency maps have been repeatedly questioned. Critiques in the literature have pointed out that many explanation methods may be misleading or unreliable, offering visual artifacts rather than faithful representations of model behavior~\cite{adebayo2018sanity, kindermans2017unreliability,longo2024explainable}. More broadly, Lipton~\cite{lipton2018mythos} argued that the goals of interpretability are often underspecified, leading to a proliferation of methods that lack a clear understanding of what constitutes a good explanation.

These concerns point to deeper, systemic challenges in how saliency explanations are conceived, evaluated, and aligned with user needs:

\textbf{Challenge 1 – Lack of Support for User-Driven Explanatory Questions.}
Users rarely ask only “Why this class?” Instead, they often seek richer, comparative justifications—such as “Why a sports car and not a convertible?” or “What distinguishes this husky from other dog breeds?”—that most conventional saliency methods are not designed to support.

\textbf{Challenge 2 – Interpretation Ambiguity from Missing Reference Frame and Granularity.}
Without explicit context about the explanation’s \textit{reference frame} (e.g., pointwise vs. contrastive) or \textit{semantic granularity} (e.g., fine-grained class vs. coarse-grained group), users may misinterpret saliency maps~\cite{VILONE202189}. A map highlighting fur texture might be meaningful for distinguishing huskies from shih-tzus but irrelevant when explaining why the image is a “dog” rather than a “cat.” This ambiguity undermines trust, interpretability, and ultimately the usability of saliency-based methods.

\textbf{Challenge 3 – Inadequate Evaluation of Contrastive and Granularity-Varying Explanations.}
Existing evaluation metrics focus almost exclusively on pointwise faithfulness—assessing how salient pixels affect the score of a single target class. They fail to capture whether an explanation meaningfully distinguishes between classes or operates coherently at different levels of semantic abstraction (e.g., vehicle vs. sports car). As a result, current benchmarks cannot validate whether an explanation aligns with the user’s intended question.

\begin{figure}[t]
\centering
\includegraphics[width=0.5\textwidth]{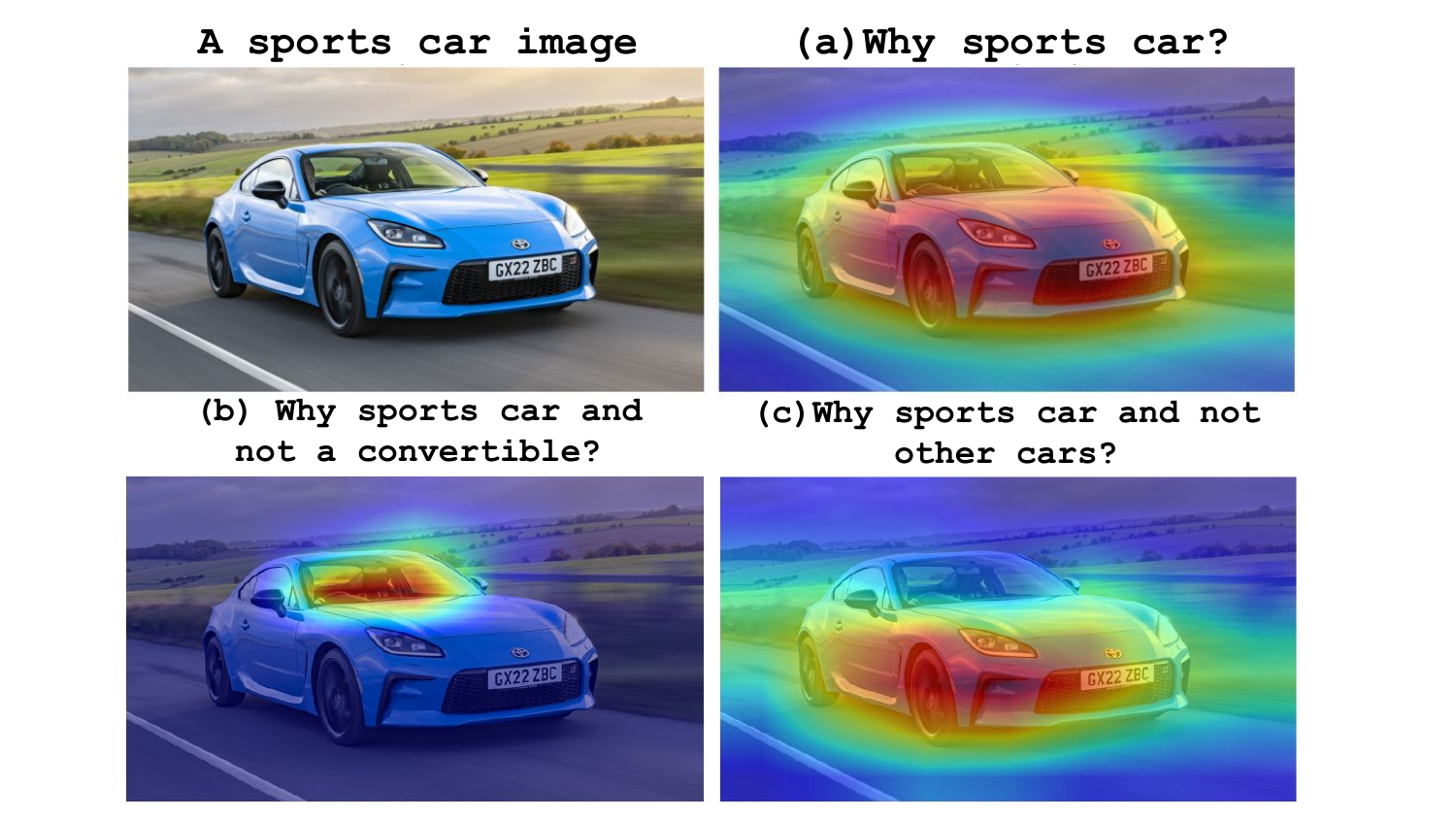}
\vspace{-4mm}
\caption{Saliency maps for a sport car using different types of questions that reflect our taxonomy: (a) Pointwise class. (b) Contrastive between two classes. (c) Contrastive between class and group.}
\vspace{-4mm}
\label{fig:xai_maps}
\end{figure}

These limitations highlight the importance of shifting attention toward the \textit{user's perspective}. What does a user actually want to know when they inspect a saliency map? Are they looking for reasons why an image was classified as a “husky,” or why it was classified as a “husky rather than a shih-tzu”? Do they care about fine-grained details or coarse-grained features? These kinds of questions suggest that saliency explanations should be considered along two important dimensions: the \textit{reference-frame} axis (pointwise vs. contrastive explanations) and the \textit{granularity} axis (fine-grained vs. coarse-grained classes). Figure~\ref{fig:xai_axis} depicts the proposed axes space, while Figure~\ref{fig:xai_maps} presents representative saliency maps that address distinct explanatory questions situated within this space.

Our contributions are as follows: First, we introduce a \textbf{novel taxonomy} of saliency explanations based on \textit{reference-frame} and \textit{granularity}, offering a conceptual structure that reflects diverse user queries. Second, we identify critical shortcomings in existing evaluation metrics, which fail to distinguish among explanation types along these axes. Third, We propose \textbf{four new faithfulness metrics} that assess explanation quality across both axes, grounded in structured perturbation and score comparison. Fourth, we contribute a \textbf{new group-level labeling for ImageNet classes}, derived from the WordNet hierarchy. This semantic grouping enables evaluation at the group granularity level. Finally, We present a comprehensive \textbf{evaluation framework}, including experiments across ten saliency methods, four model architectures, three datasets. The results provide new insights and outline foundational steps toward user-aligned, semantically meaningful explanations.
By grounding saliency evaluation in a principled, user-aligned framework, our work advances the interpretability of AI models and enables systematic, semantically coherent comparison of explanation methods.

\section{Related Work}
\label{sec:related}

\subsection{Saliency-Based Explanation Methods}
\label{subsec:related_methods}
Saliency methods form the backbone of visual explanations in computer vision, aiming to highlight the most influential regions of an input image for a model’s decision. Activation-based methods~\cite{erhan2009visualizing} use the feature-maps obtained by forward propagation in order to interpret the output prediction.
Perturbation-based methods~\cite{fong2019understanding,fong2017interpretable}
measures the output's sensitivity w.r.t. the input using random perturbations applied in the input space.
Path integration methods, such as Integrated Gradients~\cite{SundararajanTY17} integrates over the interpolated image gradients. Score-CAM~\cite{wang2020score}, a gradient-free method, generates class-specific importance maps by using activation maps as masks and measuring their effect on the target class score. With the rise of transformer-based architectures in vision, specialized explanation techniques have emerged. The rollout method~\cite{abnar2020quantifying} aggregates attention weights across layers to identify influential token relationships, while 
Transformer Attribution~\cite{chefer2021transformer} presents a class-specific Deep Taylor Decomposition method in which relevance propagation is applied for
positive and negative attributions. 
More recently, the authors introduced Generic Attention Explainability~\cite{chefer2021generic}, a generalization of Transformer Attribution for explaining Bi-Modal transformers. Iterated Integrated Attributions~\cite{Barkan_2023_ICCV} generalizes conventional path-integral methods by performing an iterated n-fold integration across multiple network layers (beyond the input) thereby measuring the n-dimensional flux through higher-order hypersurfaces. IIA quantifies how the gradient field propagates across an n-dimensional volume in the joint space of all participating network layers, including the input, and was shown to produce state-of-the-art results both for CNNs and ViTs.
Recently, contrastive methods have been introduced to provide comparative explanations, such as explaining why an image was classified as class A rather than class B~\cite{ghorbani2019towards, xie2023two, wang2023counterfactual}.
Although prior work implicitly spans multiple explanatory styles, there is a lack of a formal framework to classify and evaluate these methods systematically. We address this gap through a principled taxonomy aligned with user-centric explanation goals.

\subsection{Evaluation Metrics for Saliency Maps}
\label{subsec:related_metrics}
Faithfulness, a cornerstone concept in explanation evaluation, refers to the degree to which an explanation accurately reflects the actual decision-making process of the model~\cite{rudin2019stop}. A faithful explanation should highlight features that genuinely influence the model's prediction rather than artifacts of the explanation method itself. 
Numerous metrics have been proposed to assess saliency methods. Faithfulness-based metrics such as Insertion and Deletion~\cite{petsiuk2018rise}, AOPC~\cite{samek2017evaluating}, and MoRF/LeRF curves~\cite{samek2019towards} measure the impact of removing or adding salient pixels on the model’s prediction. Localization metrics such as the Pointing Game~\cite{zhang2018top} or bounding-box overlap~\cite{zhou2016learning} evaluate alignment between saliency maps and annotated objects. Despite their utility, most of these metrics assume a pointwise, model-centric view and fail to address contrastive reasoning or semantic granularity. For example, Insertion and Deletion evaluate influence of individual features but do not measure whether an explanation distinguishes class $A$ from a similar class $B$. Human-centered evaluations, such as trustworthiness, satisfaction, or faithfulness to user expectations, have been studied via user studies~\cite{doshi2017towards, kim2022hive}, but these are costly and often task-specific. Robustness checks such as model randomization tests~\cite{adebayo2018sanity} and input perturbation sensitivity~\cite{ghorbani2019interpretation} assess the stability or reliability of explanations.

\textbf{Limitations of Existing Contrastive Metrics.}
Recent works have proposed contrastive evaluation metrics~\cite{wang2023counterfactual,xie2023two} aimed at quantifying how well saliency maps explain model preferences between classes. While these metrics mark progress, they fail to capture key dimensions necessary for user-oriented evaluation—particularly regarding \textit{reference-frame alignment} and \textit{semantic granularity}, and \textit{correct faithfulness evaluation under structured perturbations}.
\textbf{CDS (Class Deviation Score)} measures prediction shifts after masking but lacks true contrastive framing. It evaluates only the target class $A$, without considering how feature removal affects or distinguishes a specific alternative class $B$.
\textbf{CAUC (Contrastive AUC)} evaluates whether the model remains more confident in class $A$ than class $B$ throughout the deletion process. While intuitive, the metric suffers from several limitations. It depends on a thresholding mechanism that assumes the saliency map contains many zero-valued pixels, an assumption often violated in some of the transformer-based and localized explanation methods, where saliency maps tend to be dense. This can lead to perturbations that affect only a small portion of the salient regions, resulting in unstable behavior and poor comparability across maps with different value distributions. Additionally, CAUC multiplies prediction scores during its computation, which compromises interpretability. This design choice penalizes cases where the contrastive class initially receives a low score, even if the saliency map effectively suppresses class $B$, ultimately yielding misleadingly low CAUC values for otherwise informative explanations.

\textbf{CDROP (Contrastive Drop)} extends CAUC with sparsity normalization but inherits its core limitations. 

\vspace{3pt}
In contrast, our proposed metrics evaluate saliency maps across the RFxG axes. Moreover, instead of relying on multiplication, we measure contrastiveness directly through probability differences, ensuring greater robustness and interpretability. Finally, our metrics avoids unstable thresholding mechanisms.

\subsection{Critiques of XAI and Motivation for RFxG}
\label{subsec:related_critiques}
A recurring concern in the XAI literature is that the very notion of an "explanation" remains poorly defined. \cite{lipton2018mythos} argues that interpretability is often invoked without specifying its purpose, leading to a multitude of explanation methods that address different, and sometimes conflicting, goals. This ambiguity results in an evaluation landscape that lacks coherence: it is unclear whether an explanation should be faithful to the model's internals, useful to the user, or aligned with human reasoning. Several works have attempted to address this fragmentation through taxonomic frameworks. \cite{guidotti2018survey} proposed a comprehensive taxonomy distinguishing between explanation methods based on their scope (global vs. local), model type, and transparency level. \cite{sokol2020one} further refined this by introducing the concept of "explanation profiles" that account for diverse user needs and contexts. \cite{doshi2017towards} also echo this concern, advocating for a rigorous science of interpretability grounded in clearly articulated desiderata. They propose taxonomy-based evaluation protocols to distinguish between different types of interpretability, yet most existing saliency methods are not explicitly situated within such frameworks. This lack of clarity in defining explanation goals has practical consequences. As discussed by \cite{miller2019explanation}, explanations are inherently social and user-dependent. Yet, the majority of saliency research adopts a model-centered view, optimizing visualizations without considering the user's question or context. Recent works~\cite{gilpin2018explaining,bhatt2020explainable,zhang25a} further highlight that many explanation methods fall short of being actionable or trustworthy because they are not grounded in a human-centric model of understanding. These foundational critiques motivate our proposal: saliency explanations must be designed with respect to the specific \emph{user question} they are meant to answer. We introduce a structured perspective based on the axes of \textit{reference frame} and \textit{granularity}, called \textbf{R}eference-\textbf{F}rame \textbf{x} \textbf{G}ranularity (RFxG), enabling a principled alignment between explanation methods and user intent. By introducing faithfulness-based metrics that are better suited to this framing, our approach bridges the gap between user expectations and model behavior. Our work aims to move explanations towards better correspondence between user questions and meaningful explanation evaluation for explainable AI.

\section{New Taxonomy for User-Centric Explanations}
\label{sec:taxonomy}
As outlined in Section~\ref{sec:intro}, current approaches to saliency maps exhibit several fundamental gaps. We address the first two challenges by proposing the novel Reference-Frame x Granularity (RFxG) taxonomy that decomposes attribution methods along two orthogonal axes: \textit{reference frame} and \textit{explanation granularity}. These axes create a space that helps to clarify design decisions, underlying assumptions, and downstream limitations. 

\subsection{Reference Frame: What Is the Explanation
Relative To?}
\label{sec:ref-frame}

The first axis concerns the \emph{reference frame} of the explanation, that is, what baseline or alternative hypothesis the explanation is conditioned on.\\
\textbf{Pointwise Explanations.}
Pointwise explanations aim to justify a prediction by highlighting evidence that supports the predicted class. They answer questions like: “Why did the model make this prediction for this input?” These explanations highlight regions that support the current class label without considering alternatives. Popular examples include GradCAM~\cite{selvaraju2017grad}, Guided Backpropagation~\cite{springenberg2014striving}, and Integrated Gradients~\cite{sundararajan2017axiomatic}.\\
\textbf{Contrastive Explanations.}
Contrastive explanations are defined relative to an alternative class or hypothesis. They answer questions such as: “Why did the model choose
class A over class B?”. This framing is more faithful to how humans generate explanations \cite{miller2019explanation} and helps reduce ambiguity by focusing on discriminative evidence \cite{dhurandhar2018explanations, ghorbani2019towards}. Recent works on contrastive attribution~\cite{dhurandhar2018explanations,xie2023two,wang2023counterfactual} explicitly model these differences.

\subsection{Semantic Granularity: What Level of Class
Detail is Targeted?}
\label{sec:granularity}
The second axis addresses the \emph{semantic resolution} of the explanation, are we trying to explain a broad category or a narrow one?\\
\textbf{Group-Level Explanations}
Group-level methods highlight features that are predictive of a superclass or cluster of classes. For example, when classifying an image as a \textit{husky}, a group-level explanation might emphasize general \textit{dog} features. These methods tend to generalize well and provide intuitive reasoning at a high level. Some class-agnostic or hierarchy-aware methods~\cite{ghorbani2020neuron, xie2023two} can be interpreted in this way. 
While prior methods operate implicitly at the group level, they lack a unified structured framework that links between explanation granularity and user intent. Our \textbf{RFxG} taxonomy introduces an explicit two-axis framework that distinguishes between class, group, and contrastive explanations, enabling alignment with specific user questions and guiding principled method design and evaluation.\\
\textbf{Class-Level Explanations.}
Class-level methods target fine-grained distinctions, isolating features that are specific to a single class. In the same husky example, a class-level explanation would emphasize the distinctive fur pattern or blue eyes, traits that separate huskies from other dogs. This is the dominant paradigm in the literature, as most attribution methods~\cite{selvaraju2017grad,wang2020score,sundararajan2017axiomatic} are designed for single-class specificity.
\section{Evaluating User-Perspective Saliency Maps}
\label{sec:metrics}

In what follows, we address the third challenge by introducing four novel faithfulness metrics designed to demonstrate that user-intent-driven explanations faithfully reflect the model’s behavior and decision-making process.
In this work, we shift focus toward two underexplored yet essential aspects: \textbf{contrastive metrics} and \textbf{group-level metrics}. These better align with the explanatory questions users actually ask—such as ``Why class A and not B?'' or ``What features define the entire group of cars?'' — rather than merely ``Why class A?'' Our proposed metrics are designed to measure explanation quality along the RFxG axes introduced in Sec.~\ref{sec:taxonomy}.
Our proposed metrics focus explicitly on \textbf{faithfulness}—the extent to which explanations accurately reflect the model’s true decision-making process. Faithfulness is a cornerstone of explainable AI, as it underpins user trust, enables the detection of biases, and supports effective model debugging~\cite{zhang2020explainable,fu2020fairness}. All four metrics are grounded in perturbation-based analysis, a widely adopted approach for assessing faithfulness by directly measuring how explanations influence model predictions. Let $f: \mathbb{R}^d \rightarrow \mathbb{R}^C$ be a pretrained classifier over $C$ classes, and let $x \in \mathbb{R}^d$ be an input image. Let $f_c(x)$ denote the softmax probability assigned by the classifier $f$ to class $c$ given input $x$. Let $A$ denote the predicted class and $B$ a contrastive class, while $\mathcal{G}_A$ and $\mathcal{G}_B$ denote semantic groups. An explanation method $\mathcal{E}$ produces a saliency map $M_c = \mathcal{E}(x, c)$ for class $c$. 

To enable perturbation via masking, we binarize the saliency map to obtain a top-$\alpha$ binary mask $M^{\alpha} \in \{0,1\}^d$, where 1 indicates the top $\alpha$ fraction of salient pixels (to be suppressed). We then define the perturbed image as
\begin{equation}
x_\alpha = x \odot (1 - M^{\alpha}),
\end{equation}
where $\odot$ denotes element-wise (Hadamard) multiplication, and $1$ is the all-ones vector of the same dimension as $x$. This operation zeros out the most salient regions, simulating their removal while preserving the tensor structure of the input.

Perturbation proceeds in 10\% steps from $\alpha=0.1$ to $\alpha=0.9$~\cite{chefer2021transformer}, and the \textit{Area Under the Curve (AUC)} is used to aggregate results, following established practices in perturbation-based evaluation~\cite{bach2015pixel,samek2017evaluating,petsiuk2018rise}. AUC provides a robust summary of saliency effectiveness over the full perturbation trajectory, and enables meaningful comparisons across methods by capturing both early and cumulative effects of masking. We compute all metric scores using the softmax probabilities rather than raw logits, as probabilities are bounded, interpretable, and better reflect the model’s actual output behavior. We used black pixel masking (i.e., zeroing out pixels) as the default perturbation mechanism due to its effectiveness and consistency with prior work. However, we also experimented with alternative masking strategies such as Gaussian blur, uniform noise baseline, and all other alternatives suggested in~\cite{sturmfels2020visualizing}. Across all variants, trends in performance remained consistent.
Below we present the four proposed metrics.\\
\textbf{Contrastive Contrastivity Score (CCS).}
This metric aims to assess saliency maps for contrastive questions such as \textit{``Why sports car and not convertible?''}. Given a contrastive saliency map $M_{\text{con}}^{B,A}$ that explains why class $B$ is predicted over class $A$, CCS quantifies how discriminative the identified regions are between the two classes. It computes the AUC of the prediction gap as salient regions are removed:
\begin{equation}
\text{CCS} = \mathrm{AUC}_\alpha \left[ f_A(x_\alpha) - f_B(x_\alpha) \right],
\end{equation}
where $x_\alpha = x \odot (1 - M_{\text{con}}^{B,A,\alpha})$. A high CCS indicates that removed regions are critical for distinguishing $B$ from $A$, reflecting contrastive faithfulness. In contrast to CAUC, CCS uses differences between probabilities rather than multiplication, which promotes better isolation between the terms. Moreover, it doesn't use unstable thresholds, but uses the well-established perturbation scaling approach used in~\cite{chefer2021transformer}. \\
\textbf{Class Group Contrastivity (CGC).}
CGC evaluates explanations for queries like \textit{``Why sports car and not other cars?''}. Here, the contrastive saliency map $M_{\text{con}}^{A,\mathcal{G}_A}$ highlights features distinguishing class $A$ from others in its semantic group $\mathcal{G}_A$. CGC aggregates the drop in $f_A$ and rise in confidences of competing group members after masking:
\begin{multline}
\text{CGC} = \mathrm{AUC}_\alpha \left[ \frac{1}{2} \left( \frac{1}{|\mathcal{G}_A|} \sum_{k \in \mathcal{G}_A} (f_k(x_\alpha) - f_k(x)) \right. \right. \\
\left. \left. + (f_A(x) - f_A(x_\alpha)) \right) \right],
\end{multline}
where $x_\alpha = x \odot (1 - M_{\text{con}}^{A,\mathcal{G}_A,\alpha})$. CGC evaluates inter-group discriminativeness, essential in fine-grained tasks.\\
\textbf{Pointwise Group Score (PGS).}
PGS evaluates explanations for questions such as \textit{``Why Car?''}, where the explanation pertains to an entire group. Given a pointwise group-level saliency map $M_{\text{pt}}^{\mathcal{G}_A}$, it computes the average confidence drop across the group when salient regions are removed:
\begin{equation}
\text{PGS} = \mathrm{AUC}_\alpha \left[ \frac{1}{|\mathcal{G}_A|} \sum_{k \in \mathcal{G}_A} \left( f_k(x) - f_k(x \odot (1 - M_{\text{pt}}^{\mathcal{G}_A,\alpha})) \right) \right].
\end{equation}
PGS thus captures semantic generality rather than specificity, distinguishing it from class-centric faithfulness metrics.\\
\textbf{Contrastive Group Score (CGS).}
This metric measures contrastivity between groups, evaluating saliency maps for questions like \textit{``Why Car and not Truck?''}. The saliency map $M_{\text{con}}^{\mathcal{G}_A,\mathcal{G}_B}$ captures features discriminative between two semantic groups. CGS measures whether deletion of those features lowers $\mathcal{G}_A$ confidence while raising $\mathcal{G}_B$:
\begin{multline}
\text{CGS} = 
\mathrm{AUC}_\alpha \left[ \frac{1}{2} \left( \frac{1}{|\mathcal{G}_A|} \sum_{k \in \mathcal{G}_A} (f_k(x) - f_k(x_\alpha)) \right. \right. \\
\left. \left. + \frac{1}{|\mathcal{G}_B|} \sum_{j \in \mathcal{G}_B} (f_j(x_\alpha) - f_j(x)) \right) \right],
\end{multline}
where $x_\alpha = x \odot (1 - M_{\text{con}}^{\mathcal{G}_A,\mathcal{G}_B,\alpha})$. This metric captures both suppression and promotion across groups.

\vspace{1em}
Together, CCS, CGC, PGS, and CGS form a comprehensive, user-aligned suite of evaluation metrics for RFxG's saliency maps. They provide theoretical grounding for contrastive and group-based faithfulness—two areas neglected in existing evaluation protocols. These metrics support a more nuanced, purpose-driven understanding of visual explanations and complement, rather than replace, standard metrics.

\section{Evaluation Framework and Experiments}
\label{sec:experiments}
In this section, we present a comprehensive experimental framework to evaluate the effectiveness of our proposed RFxG user-centric taxonomy and associated metrics. 
The experiments were conducted on an NVIDIA DGX 8xA100 server, utilizing the Pytorch package.
We note that the reported quantitative comparative results in this paper are statistically significant, as determined by a t-test with a p-value of 0.05. Due to space constraints, additional technical and implementation details, including extended analyses and results, are provided in the Appendix.

\subsection{Datasets}
We construct our benchmark using the validation sets of three widely adopted datasets: (1) PASCAL Visual Object Classes (\textbf{VOC})~\cite{Everingham2009ThePV}
(2) ImageNet ILSVRC 2012 (\textbf{IN})~\cite{imagenet}.
(3) Microsoft Common Objects in COntext 2017 (\textbf{COCO})~\cite{lin2014microsoft}.
In our evaluation, the highest scoring classes are considered as the ground-truth.
Semantic groups were constructed using the WordNet hierarchy\footnote{\url{https://wordnet.princeton.edu/}}. For each class, we identified candidate superordinate concepts in the WordNet graph that contain at least five subordinate leaf nodes. Further details provided in the Appendix. For example, \textit{sports car}, \textit{cab}, and \textit{limousine} were grouped under \textit{car}. This group-to-class ontology enabled coarse-to-fine explanations along our granularity axis. The resulting dataset of image-class-group triplets will be publicly released as a benchmark for group and contrast-based explanation research.
\vspace{-1mm}
\subsection{Models and Explanation Methods}
We evaluated four image classification models: ResNet-50 (\textbf{RN})~\cite{he2016resnet}, ConvNext-Base (\textbf{CN})~\cite{Liu2022ACF}, ViT-Base (\textbf{ViT-B}) and ViT-Small (\textbf{ViT-S})~\cite{dosovitskiy2020image}. For \textbf{CNNs}, we applied five widely used explanation methods: Grad-CAM (\textbf{GC})~\cite{selvaraju2017grad}, Integrated Gradients (\textbf{IG})~\cite{sundararajan2017axiomatic}, Score-CAM (\textbf{SC})~\cite{wang2020score}, SHAP~\cite{lundberg2017unified}, and Integrated Iterated Attributions (\textbf{IIA})~\cite{Barkan_2023_ICCV}. For \textbf{Transformers}, we used: Grad-CAM-ViT (\textbf{GCV})~\cite{chefer2021transformer}, Attention Rollout (\textbf{Rollout})~\cite{abnar2020quantifying}, Generic Attention-model Explainability (\textbf{GAE})~\cite{chefer2021generic}, Transformer Attribution (\textbf{TAttr})~\cite{chefer2021transformer}, and IIA. Our experiments focus on saliency methods that produce a single map per question, and are agnostic to model internals, excluding concept-based and generative explanation techniques. To enable consistent evaluation across the full RFxG taxonomy, we adapted a subset of explanation methods - leveraging publicly available implementations where possible~\cite{cntrstv_adapt_nuerips22,eriksson2025reproducibility}, and applying minimal modifications when necessary—to align with the explanatory goals of each evaluation setting. For instance, we used Contrastive-GradCAM~\cite{cntrstv_adapt_nuerips22} to adapt Grad-CAM for contrastive explanations, and implemented a minimally modified version of SHAP, referred to as Contrastive-SHAP, to support RF×G-oriented evaluation. All methods were evaluated using their default hyperparameters. Further implementation details are provided in the Appendix.

\vspace{-1mm}
\subsection{Metric Computation and Evaluation Setup}
For each image, we computed saliency maps and metric scores using the following selection criteria: (1) \textbf{Class A}: top predicted class by the model.
(2) \textbf{Class B}: the second-highest-scoring class from the same semantic group as A.
(3) \textbf{Group $\mathcal{G}_A$ (pointwise)}: defined using WordNet groupings.
(4) \textbf{Group $\mathcal{G}_A$ (contrastive)}: all classes in the group excluding A.
This strategy ensured meaningful contrastive setups and valid group-level comparisons. All metrics followed the masking protocol using black pixels.

\subsection{Quantitative Results}
\begin{table}[t!]
\centering
\caption{Evaluation results across datasets, models, and explanation methods. For every metric in the table, the results were provided using a method suitable for the metric's purpose. 
}
\label{tab:results}
\vspace{-2mm}
\scalebox{0.75}{
\begin{tabular}{lll|cccc}
\toprule
Dataset & Model & Method & CCS $\uparrow$ & CGC $\uparrow$ & PGS $\uparrow$ & CGS $\uparrow$ \\
\midrule
\multirow{20}{*}{COCO}
& \multirow{5}{*}{CN} & IIA & \textbf{25.20} & \textbf{5.25} & \textbf{46.57} & \textbf{36.38} \\
& & GC & \underline{22.48} & \underline{4.67} & \underline{41.00} & \underline{32.33} \\
& & SC & 22.21 & 4.49 & 41.35 & 31.96 \\
& & SHAP & 16.51 & 3.33 & 32.81 & 26.61 \\
& & IG & 13.73 & 3.02 & 27.68 & 23.21 \\
\cmidrule(lr){2-7}
& \multirow{5}{*}{RN} & IIA & \textbf{24.98} & \textbf{5.14} & \textbf{46.45} & \textbf{35.78} \\
& & GC & \underline{21.86} & \underline{4.63} & \underline{42.36} & \underline{32.30} \\
& & SC & 21.41 & 4.34 & 41.48 & 31.88 \\
& & SHAP & 16.10 & 3.36 & 33.04 & 25.74 \\
& & IG & 13.48 & 2.89 & 28.33 & 23.45 \\
\cmidrule(lr){2-7}
& \multirow{5}{*}{ViT-B} & IIA & \textbf{19.69} & \textbf{4.56} & \textbf{38.92} & \textbf{29.78} \\
& & TAttr & \underline{17.81} & \underline{4.18} & \underline{36.62} & \underline{27.79} \\
& & GAE & 17.58 & 4.07 & 35.99 & 27.39 \\
& & Rollout & 14.30 & 3.24 & 29.05 & 22.97 \\
& & GCV & 12.59 & 2.74 & 27.16 & 20.58 \\
\cmidrule(lr){2-7}
& \multirow{5}{*}{ViT-S} & IIA & \textbf{20.01} & \textbf{4.50} & \textbf{39.50} & \textbf{29.15} \\
& & TAttr & \underline{17.79} & \underline{4.21} & \underline{37.23} & \underline{27.55} \\
& & GAE & 17.51 & 4.06 & 36.67 & 27.15 \\
& & Rollout & 14.23 & 3.25 & 29.70 & 22.53 \\
& & GCV & 12.61 & 2.66 & 26.29 & 20.95 \\
\midrule
\multirow{20}{*}{IN}
& \multirow{5}{*}{CN} & IIA & \textbf{25.15} & \textbf{5.32} & \textbf{46.03} & \textbf{36.56} \\
& & GC & \underline{22.72} & \underline{4.58} & \underline{42.70} & \underline{33.78} \\
& & SC & 22.29 & 4.44 & 41.97 & 33.05 \\
& & SHAP & 16.76 & 3.40 & 32.26 & 26.04 \\
& & IG & 13.57 & 2.61 & 27.62 & 23.58 \\
\cmidrule(lr){2-7}
& \multirow{5}{*}{RN} & IIA & \textbf{25.07} & \textbf{5.38} & \textbf{46.17} & \textbf{36.42} \\
& & GC & \underline{22.64} & \underline{4.52} & \underline{42.58} & \underline{32.51} \\
& & SC & 22.10 & 4.40 & 41.66 & 31.66 \\
& & SHAP & 15.91 & 3.32 & 33.16 & 26.90 \\
& & IG & 13.48 & 2.68 & 27.64 & 23.53 \\
\cmidrule(lr){2-7}
& \multirow{5}{*}{ViT-B} & IIA & \textbf{20.12} & \textbf{4.51} & \textbf{39.24} & \textbf{29.76} \\
& & TAttr & \underline{18.47} & \underline{4.07} & \underline{37.21} & \underline{27.13} \\
& & GAE & 18.24 & 3.99 & 36.59 & 26.71 \\
& & Rollout & 14.66 & 3.15 & 30.20 & 22.59 \\
& & GCV & 12.49 & 2.70 & 26.81 & 20.63 \\
\cmidrule(lr){2-7}
& \multirow{5}{*}{ViT-S} & IIA & \textbf{19.86} & \textbf{4.63} & \textbf{39.31} & \textbf{29.91} \\
& & TAttr & \underline{18.45} & \underline{4.02} & \underline{35.71} & \underline{27.49} \\
& & GAE & 18.17 & 3.94 & 35.14 & 27.10 \\
& & Rollout & 14.42 & 3.22 & 29.16 & 23.04 \\
& & GCV & 12.68 & 2.65 & 26.29 & 20.52 \\
\bottomrule
\end{tabular}}
\vspace{-3mm}
\end{table}
\begin{table*}[ht]
\centering
\caption{Comparison of explanation methods using Deletion, CAUC ($\times 10^{-3}$), CDROP ($\times 10^{-2}$), and CDS. Best values are \textbf{bolded}, second-best are \underline{underlined}.}
\label{tab:results_table}
\vspace{-2mm}
\scalebox{0.65}{
\begin{tabular}{ll|cccc|cccc|cccc}
\toprule
& & \multicolumn{4}{c|}{\textbf{COCO}} & \multicolumn{4}{c|}{\textbf{IN}} & \multicolumn{4}{c}{\textbf{VOC}} \\
Model & Method & Deletion$\downarrow$ & CAUC$\downarrow$ & CDROP$\uparrow$ & CDS$\uparrow$ & Deletion$\downarrow$ & CAUC$\downarrow$ & CDROP$\uparrow$ & CDS$\uparrow$ & Deletion$\downarrow$ & CAUC$\downarrow$ & CDROP$\uparrow$ & CDS$\uparrow$ \\
\midrule
\multirow{5}{*}{RN}
& IIA & \textbf{11.24} & \textbf{3.09} & 7.88 & \textbf{6.29} & \textbf{11.41} & \textbf{3.16} & 7.85 & \textbf{6.24} & \textbf{11.35} & \textbf{3.13} & 7.68 & \textbf{6.21} \\
& SC & \underline{14.31} & \underline{3.14} & \textbf{8.36} & \underline{6.18} & \underline{14.59} & \underline{3.22} & \textbf{8.40} & \underline{6.11} & \underline{14.45} & \underline{3.17} & \textbf{8.32} & \underline{6.12} \\
& GC & 14.87 & 3.21 & \underline{8.11} & 6.04 & 14.73 & 3.29 & \underline{8.15} & 6.02 & 14.62 & 3.22 & \underline{8.08} & 6.01 \\
& IG & 18.22 & 3.30 & 7.45 & 5.92 & 18.45 & 3.35 & 7.54 & 5.95 & 18.33 & 3.30 & 7.50 & 5.89 \\
& SHAP & 20.19 & 3.51 & 5.63 & 5.60 & 20.29 & 3.46 & 5.63 & 5.59 & 20.12 & 3.48 & 5.61 & 5.52 \\
\midrule
\multirow{5}{*}{CN}
& IIA & \textbf{11.53} & \textbf{3.12} & 7.91 & \textbf{6.27} & \textbf{11.69} & \textbf{3.18} & 7.70 & \textbf{6.20} & \textbf{11.62} & \textbf{3.15} & 7.83 & \textbf{6.19} \\
& SC & \underline{14.67} & \underline{3.16} & \textbf{8.31} & \underline{6.13} & \underline{14.78} & \underline{3.25} & \textbf{8.35} & \underline{6.09} & \underline{14.72} & \underline{3.19} & \textbf{8.29} & \underline{6.08} \\
& GC & 14.98 & 3.25 & \underline{8.10} & 6.00 & 14.95 & 3.29 & \underline{8.11} & 6.08 & 14.91 & 3.26 & \underline{8.07} & 6.00 \\
& IG & 18.51 & 3.33 & 7.38 & 5.91 & 18.64 & 3.37 & 7.42 & 5.88 & 18.59 & 3.35 & 7.41 & 5.87 \\
& SHAP & 20.25 & 3.52 & 5.59 & 5.54 & 20.31 & 3.48 & 5.52 & 5.50 & 20.28 & 3.50 & 5.54 & 5.51 \\
\bottomrule
\end{tabular}}
\vspace{-3mm}
\end{table*}
Table~\ref{tab:results} presents results for CNN and ViT models across three datasets using our proposed RFxG metrics. VOC results are included in the Appendix due to space constraints. Table~\ref{tab:results_table} reports Deletion, CAUC, CDROP, and CDS scores for CNNs across all datasets, provided for comparison with RFxG metrics. Our quantitative evaluation, shown in Table~\ref{tab:results}. The results reveal several compelling insights that underscore the need for explanation methods aligned with contrastive and semantic granularity axes. Across all datasets and model architectures, IIA consistently outperforms the other methods, particularly on the contrastive-class (CCS) and group-level (PGS, CGS) metrics. This aligns with IIA’s unique design that integrates attributions over multiple intermediate network layers~\cite{Barkan_2023_ICCV}. By leveraging multi-level features at various semantic scales, IIA captures both fine-grained and high-level evidence, an essential thing for explaining subtle class distinctions and inter-group variability. Furthermore, the more focused and spatially coherent maps produced by IIA allow it to better isolate discriminative cues, especially in contrastive settings.
Notably, \textbf{performance is consistently higher on PGS and CGS} than on CGC and CCS, across nearly all methods and models. This discrepancy suggests that generating faithful explanations for group-level concepts is easier than for individual class distinctions. The challenge of explaining a class \textit{within} a semantically coherent group (as in CGC) is inherently harder, likely due to the shared low and mid-level features between sibling classes. Conversely, contrastive group questions like ``Why Car and not Truck?'' (CGS) span broader conceptual boundaries and offer more readily distinguishable patterns for saliency methods to exploit. Interestingly, CCS scores still exceed CGC, implying that contrasting a class against a specific alternative may be easier than contrasting it against its entire group, possibly due to the dilution of discriminative power across many similar classes.
For Transformer-based models, TAttr ranks consistently second after IIA across all metrics. The ability of TAttr to combine attention weights with layer-wise relevance propagation allows it to capture important specific cues. Its improved performance compared to methods like Rollout or GCV also demonstrates the value of explicitly modeling gradient flow and relevance.
IG underperforms across all metrics, particularly in contrastive settings. As also noted in prior work~\cite{Barkan_2023_ICCV}, IG tends to produce coarse, diffuse maps that highlight many non-discriminative regions. These scattered attributions hinder its ability to isolate features relevant for explaining class uniqueness or group separation, reinforcing the necessity for explanation methods with higher spatial precision and stronger semantic selectivity.
Surprisingly, despite its overall superiority, IIA ranks \textbf{only third on CDROP}. This result is inconsistent with its focused and contrastive visual behavior. This suggests that existing contrastive metrics like CDROP may fail to fully reflect the discriminative power of focused maps, particularly when penalizing sparse attributions. This discrepancy supports our motivation for introducing CCS and CGC as more robust and theoretically grounded contrastive metrics that avoid artifacts from density normalization and thresholding, as seen in CAUC and CDROP~\cite{xie2023two}.
Lastly, we emphasize that the consistently strong results achieved on the PGS and CGS metrics indicate the feasibility of generating high-quality group-based explanations. 
This capability represents a promising direction in XAI research, enabling users to query explanations not only for class-specific decisions but also for semantic abstractions at the group level. Such capability is especially important in domains like medical imaging and autonomous driving, where higher-level categories carry more practical relevance. 
Overall, our results validate both the necessity and effectiveness of our proposed evaluation framework and RFxG taxonomy. They expose key differences in how explanation methods behave across the axes of reference-frame and granularity, confirming that current pointwise-centric metrics are insufficient. 

\subsection{Qualitative Results}
Figure~\ref{fig:xai_maps} presents saliency maps for a sports car input image for the following questions: (a) "Why sports car?" (Pointwise class) - using the original GC.
(b) "Why sports car and not a convertible?" (Contrastive between two classes) - using Class-Contrastive-GradCAM~\cite{cntrstv_adapt_nuerips22}. (c) "Why sports car and not
other cars?" (Contrastive between
class and group) - using Class-Group-Contrastive-GradCAM. All these saliency maps were provided using the RN model.
Implementation details for Class-Contrastive-GradCAM and Class-Group-Contrastive-GradCAM are provided in the Appendix.
Finally, additional qualitative results can be found in the Appendix.

\section{Conclusion}
\label{sec:conclusion}

We resolve the core ambiguity in saliency map interpretation by introducing the Reference-Frame × Granularity (RFxG) taxonomy, a dual-axis framework differentiating pointwise versus contrastive explanations and fine- versus coarse-grained semantics. This user-intent-driven perspective addresses the fundamental question: “What do saliency maps represent?” We expose critical gaps in current evaluation metrics, which largely emphasize pointwise faithfulness while neglecting contrastive reasoning and semantic granularity. To address this, we propose four novel metrics assessing explanation quality across both RFxG dimensions. Comprehensive evaluation over ten methods, four architectures, and three datasets reveals these limitations and demonstrates IIA’s consistent superiority in capturing contrastive evidence and semantic groupings. By validating semantic groupings and demonstrating practical utility, our framework provides a rigorous foundation for producing explanations that are both faithful to models and aligned with user needs. Further limitations and future directions are discussed in the Appendix.

\bibliography{aaai2026}

@String(CVPR  = {IEEE Conf. Comput. Vis. Pattern Recog.})

@String(ICCV  = {Int. Conf. Comput. Vis.})

@String(NeurIPS = {Adv. Neural Inform. Process. Syst.})

@String(ICML  = {Int. Conf. Mach. Learn.})

@String(AAAI  = {AAAI})

@String(CVPR  = {CVPR})

@String(ICCV  = {ICCV})

@String(NeurIPS = {NeurIPS})

@String(ICML  = {ICML})

@String{Computing = "Computing" }

@String{Computer = "{IEEE} Computer" }

@String{Springer = "Springer-Verlag" }

@ArtifactSoftware{R,
    title = {R: A Language and Environment for Statistical Computing},
    author = {{R Core Team}},
    organization = {R Foundation for Statistical Computing},
    address = {Vienna, Austria},
    year = {2019},
    url = {https://www.R-project.org/},
}

@String(CVPR= {IEEE Conf. Comput. Vis. Pattern Recog.})

@String(ICCV= {Int. Conf. Comput. Vis.})

@String(AAAI = {AAAI})

@article{Everingham2009ThePV,
  title={The Pascal Visual Object Classes (VOC) Challenge},
  author={Mark Everingham and Luc Van Gool and Christopher K. I. Williams and John M. Winn and Andrew Zisserman},
  journal={International Journal of Computer Vision},
  year={2009},
  volume={88},
  pages={303-338}
}

@inproceedings{SundararajanTY17,
  author    = {Mukund Sundararajan and
               Ankur Taly and
               Qiqi Yan},
  title     = {Axiomatic Attribution for Deep Networks},
  booktitle = {Proceedings of the 34th International Conference on Machine Learning,
               {ICML} 2017},
  pages     = {3319--3328},
  year      = {2017}
}

@inproceedings{selvaraju2017grad,
  title={Grad-CAM: Visual explanations from deep networks via gradient-based localization},
  author={Selvaraju, Ramprasaath R and Cogswell, Michael and Das, Abhishek and Vedantam, Ramakrishna and Parikh, Devi and Batra, Dhruv},
  booktitle={Proceedings of the IEEE international conference on computer vision},
  pages={618--626},
  year={2017}
}

@inproceedings{lundberg2017unified,
  title={A unified approach to interpreting model predictions},
  author={Lundberg, Scott M and Lee, Su-In},
  booktitle={Advances in Neural Information Processing Systems},
  pages={4765--4774},
  year={2017}
}

@article{erhan2009visualizing,
  title={Visualizing higher-layer features of a deep network},
  author={Erhan, Dumitru and Bengio, Yoshua and Courville, Aaron and Vincent, Pascal},
  journal={University of Montreal},
  volume={1341},
  number={3},
  pages={1},
  year={2009}
}

@article{smilkov2017smoothgrad,
  title={Smoothgrad: removing noise by adding noise},
  author={Smilkov, Daniel and Thorat, Nikhil and Kim, Been and Vi{\'e}gas, Fernanda and Wattenberg, Martin},
  journal={arXiv preprint arXiv:1706.03825},
  year={2017}
}

@article{Liu2022ACF,
  title={A ConvNet for the 2020s},
  author={Zhuang Liu and Hanzi Mao and Chaozheng Wu and Christoph Feichtenhofer and Trevor Darrell and Saining Xie},
  journal={2022 IEEE/CVF Conference on Computer Vision and Pattern Recognition (CVPR)},
  year={2022},
  pages={11966-11976}
}

@article{guidotti2018survey,
  title={A survey of methods for explaining black box models},
  author={Guidotti, Riccardo and Monreale, Anna and Ruggieri, Salvatore and Turini, Franco and Giannotti, Fosca and Pedreschi, Dino},
  journal={ACM computing surveys (CSUR)},
  volume={51},
  number={5},
  pages={1--42},
  year={2018},
  publisher={ACM New York, NY, USA}
}

@article{abnar2020quantifying,
    IDS = {rollout},
  title={Quantifying Attention Flow in Transformers},
  author={Abnar, Samira and Zuidema, Willem},
  journal={arXiv preprint arXiv:2005.00928},
  year={2020}
}

@inproceedings{he2016resnet,
  title={Deep residual learning for image recognition},
  author={He, Kaiming and Zhang, Xiangyu and Ren, Shaoqing and Sun, Jian},
  booktitle={Proceedings of the IEEE Conference on Computer Vision and Pattern Recognition},
  pages={770--778},
  year={2016}
}

@article{springenberg2014striving,
  title={Striving for simplicity: The all convolutional net},
  author={Springenberg, Jost Tobias and Dosovitskiy, Alexey and Brox, Thomas and Riedmiller, Martin},
  journal={arXiv preprint arXiv:1412.6806},
  year={2014}
}

@inproceedings{zhou2016learning,
  title={Learning deep features for discriminative localization},
  author={Zhou, Bolei and Khosla, Aditya and Lapedriza, Agata and Oliva, Aude and Torralba, Antonio},
  booktitle={Proceedings of the IEEE conference on computer vision and pattern recognition},
  pages={2921--2929},
  year={2016}
}

@article{samek2017evaluating,
  title={Evaluating the visualization of what a deep neural network has learned},
  author={Samek, Wojciech and Binder, Alexander and Montavon, Gr{\'e}goire and Lapuschkin, Sebastian and M{\"u}ller, Klaus-Robert},
  journal={IEEE transactions on neural networks and learning systems},
  volume={28},
  number={11},
  pages={2660--2673},
  year={2017},
  publisher={IEEE}
}

@inproceedings{sundararajan2017axiomatic,
  title={Axiomatic attribution for deep networks},
  author={Sundararajan, Mukund and Taly, Ankur and Yan, Qiqi},
  booktitle={Proceedings of the 34th International Conference on Machine Learning-Volume 70},
  pages={3319--3328},
  year={2017},
  organization={JMLR. org}
}

@inproceedings{adebayo2018sanity,
  title={Sanity checks for saliency maps},
  author={Adebayo, Julius and Gilmer, Justin and Muelly, Michael and Goodfellow, Ian and Hardt, Moritz and Kim, Been},
  booktitle={Advances in Neural Information Processing Systems},
  pages={9505--9515},
  year={2018}
}

@article{zhang2018top,
  title={Top-down neural attention by excitation backprop},
  author={Zhang, Jianming and Bargal, Sarah Adel and Lin, Zhe and Brandt, Jonathan and Shen, Xiaohui and Sclaroff, Stan},
  journal={International Journal of Computer Vision},
  volume={126},
  number={10},
  pages={1084--1102},
  year={2018},
  publisher={Springer}
}

@inproceedings{fong2019understanding,
  title={Understanding deep networks via extremal perturbations and smooth masks},
  author={Fong, Ruth and Patrick, Mandela and Vedaldi, Andrea},
  booktitle={Proceedings of the IEEE International Conference on Computer Vision},
  pages={2950--2958},
  year={2019}
}

@inproceedings{fong2017interpretable,
  title={Interpretable explanations of black boxes by meaningful perturbation},
  author={Fong, Ruth C and Vedaldi, Andrea},
  booktitle={Proceedings of the IEEE International Conference on Computer Vision},
  pages={3429--3437},
  year={2017}
}

@InProceedings{imagenet,
  author =    {Deng, J. and Dong, W. and Socher, R. and Li, L.-J. and Li, K. and Fei-Fei, L.},
  title =     {{ImageNet: A Large-Scale Hierarchical Image Database}},
  booktitle = {Computer Vision and Pattern Recognition (CVPR)},
  year =      {2009},
  bibsource = {http://www.image-net.org/papers/imagenet_cvpr09.bib}
}

@inproceedings{chefer2021transformer,
  title={Transformer interpretability beyond attention visualization},
  author={Chefer, Hila and Gur, Shir and Wolf, Lior},
  booktitle={Proceedings of the IEEE/CVF Conference on Computer Vision and Pattern Recognition},
  pages={782--791},
  year={2021}
}

@article{dosovitskiy2020image,
  title={An image is worth 16x16 words: Transformers for image recognition at scale},
  author={Dosovitskiy, Alexey and Beyer, Lucas and Kolesnikov, Alexander and Weissenborn, Dirk and Zhai, Xiaohua and Unterthiner, Thomas and Dehghani, Mostafa and Minderer, Matthias and Heigold, Georg and Gelly, Sylvain and others},
  journal={arXiv preprint arXiv:2010.11929},
  year={2020}
}

@article{petsiuk2018rise,
  title={Rise: Randomized input sampling for explanation of black-box models},
  author={Petsiuk, Vitali and Das, Abir and Saenko, Kate},
  journal={arXiv preprint arXiv:1806.07421},
  year={2018}
}

@inproceedings{chefer2021generic,
  title={Generic attention-model explainability for interpreting bi-modal and encoder-decoder transformers},
  author={Chefer, Hila and Gur, Shir and Wolf, Lior},
  booktitle={Proceedings of the IEEE/CVF International Conference on Computer Vision},
  pages={397--406},
  year={2021}
}

@inproceedings{wang2020score,
  title={Score-CAM: Score-weighted visual explanations for convolutional neural networks},
  author={Wang, Haofan and Wang, Zifan and Du, Mengnan and Yang, Fan and Zhang, Zijian and Ding, Sirui and Mardziel, Piotr and Hu, Xia},
  booktitle={Proceedings of the IEEE/CVF conference on computer vision and pattern recognition workshops},
  pages={24--25},
  year={2020}
}

@misc{lin2014microsoft,
  author = {Lin, Tsung-Yi and Maire, Michael and Belongie, Serge and Bourdev, Lubomir and Girshick, Ross and Hays, James and Perona, Pietro and Ramanan, Deva and Zitnick, C. Lawrence and Dollár, Piotr},
  description = {Microsoft COCO: Common Objects in Context},
  title = {Microsoft COCO: Common Objects in Context},
  url = {http://arxiv.org/abs/1405.0312},
  year = 2014
}

@inproceedings{barkan2021grad,
  title={Grad-sam: Explaining transformers via gradient self-attention maps},
  author={Barkan, Oren and Hauon, Edan and Caciularu, Avi and Katz, Ori and Malkiel, Itzik and Armstrong, Omri and Koenigstein, Noam},
  booktitle={Proceedings of the 30th ACM International Conference on Information \& Knowledge Management},
  pages={2882--2887},
  year={2021}
}

@inproceedings{barkan2020explainable,
  title={Explainable recommendations via attentive multi-persona collaborative filtering},
  author={Barkan, Oren and Fuchs, Yonatan and Caciularu, Avi and Koenigstein, Noam},
  booktitle={Proceedings of the 14th ACM Conference on Recommender Systems},
  pages={468--473},
  year={2020}
}

@article{sturmfels2020visualizing,
  author = {Sturmfels, Pascal and Lundberg, Scott and Lee, Su-In},
  title = {Visualizing the Impact of Feature Attribution Baselines},
  journal = {Distill},
  year = {2020},
  note = {https://distill.pub/2020/attribution-baselines},
  doi = {10.23915/distill.00022}
}

@inproceedings{barkan2021gam,
  title={GAM: Explainable Visual Similarity and Classification via Gradient Activation Maps},
  author={Barkan, Oren and Armstrong, Omri and Hertz, Amir and Caciularu, Avi and Katz, Ori and Malkiel, Itzik and Koenigstein, Noam},
  booktitle={Proceedings of the 30th ACM International Conference on Information \& Knowledge Management},
  pages={68--77},
  year={2021}
}

@InProceedings{Barkan_2023_ICCV,
    author    = {Barkan, Oren and Elisha, Yehonatan and Asher, Yuval and Eshel, Amit and Koenigstein, Noam},
    title     = {Visual Explanations via Iterated Integrated Attributions},
    booktitle = {Proceedings of the IEEE/CVF International Conference on Computer Vision (ICCV)},
    month     = {October},
    year      = {2023},
    pages     = {2073-2084}
}

@inproceedings{barkan2024learning,
  title={A Learning-based Approach for Explaining Language Models},
  author={Barkan, Oren and Toib, Yonatan and Elisha, Yehonatan and Koenigstein, Noam},
  booktitle={Proceedings of the 33rd ACM International Conference on Information and Knowledge Management},
  pages={98--108},
  year={2024}
}

@inproceedings{barkan2024llm,
  title={LLM Explainability via Attributive Masking Learning},
  author={Barkan, Oren and Toib, Yonatan and Elisha, Yehonatan and Weill, Jonathan and Koenigstein, Noam},
  booktitle={Findings of the Association for Computational Linguistics: EMNLP 2024},
  pages={9522--9537},
  year={2024}
}

@inproceedings{barkan2025bee,
  title={BEE: Metric-Adapted Explanations via Baseline Exploration-Exploitation},
  author={Barkan, Oren and Elisha, Yehonatan and Weill, Jonathan and Koenigstein, Noam},
  booktitle={Proceedings of the AAAI Conference on Artificial Intelligence},
  volume={39},
  number={2},
  pages={1835--1843},
  year={2025}
}

@inproceedings{barkan2024improving,
  title={Improving LLM Attributions with Randomized Path-Integration},
  author={Barkan, Oren and Elisha, Yehonatan and Toib, Yonatan and Weill, Jonathan and Koenigstein, Noam},
  booktitle={Findings of the Association for Computational Linguistics: EMNLP 2024},
  pages={9430--9446},
  year={2024}
}

@inproceedings{haddad2025soft,
  title={Soft Local Completeness: Rethinking Completeness in XAI},
  author={Haddad, Ziv Weiss and Barkan, Oren and Elisha, Yehonatan and Koenigstein, Noam},
  booktitle={Proceedings of the IEEE/CVF International Conference on Computer Vision},
  pages={19794--19804},
  year={2025}
}

@inproceedings{ghorbani2019towards,
  title={Towards automatic concept-based explanations},
  author={Ghorbani, Amirata and Wexler, James and Zou, James and Kim, Been},
  booktitle={Advances in Neural Information Processing Systems (NeurIPS)},
  year={2019}
}

@article{kindermans2017unreliability,
  title={The (un) reliability of saliency methods},
  author={Kindermans, Pieter-Jan and Hooker, Sara and Adebayo, Julius and Alber, Maximilian and Sch{\"u}tt, Kristof T and D{\"a}hne, Sven and Erhan, Dumitru and Kim, Been},
  journal={arXiv preprint arXiv:1711.00867},
  year={2017}
}

@article{lipton2018mythos,
  title={The mythos of model interpretability},
  author={Lipton, Zachary C},
  journal={Queue},
  volume={16},
  number={3},
  pages={31--57},
  year={2018},
  publisher={ACM}
}

@inproceedings{doshi2017towards,
  title={Towards a rigorous science of interpretable machine learning},
  author={Doshi-Velez, Finale and Kim, Been},
  journal={arXiv preprint arXiv:1702.08608},
  year={2017}
}

@inproceedings{ghorbani2019interpretation,
  title={Interpretation of neural networks is fragile},
  author={Ghorbani, Amirata and Abid, Abubakar and Zou, James Y},
  booktitle={Proceedings of the AAAI Conference on Artificial Intelligence},
  year={2019}
}

@article{miller2019explanation,
  title={Explanation in artificial intelligence: Insights from the social sciences},
  author={Miller, Tim},
  journal={Artificial Intelligence},
  volume={267},
  pages={1--38},
  year={2019}
}

@article{bhatt2020explainable,
  title={Explainable machine learning in deployment},
  author={Bhatt, Umang and Xiang, Alice and Sharma, Shubham and Weller, Adrian and Taly, Ankur and Jia, Yoni and Ghosh, Shubham and Yona, Gal and Mirel, Daniel and others},
  journal={Proceedings of the 2020 Conference on Fairness, Accountability, and Transparency (FAT* ’20)},
  year={2020}
}

@inproceedings{gilpin2018explaining,
  title={Explaining explanations: An overview of interpretability of machine learning},
  author={Gilpin, Leilani H and Bau, David and Zoran, Daniel and Bajwa, Ayesha and Specter, Michael and Kagal, Lalana},
  booktitle={2018 IEEE 5th International Conference on data science and advanced analytics (DSAA)},
  pages={80--89},
  year={2018},
  organization={IEEE}
}

@inproceedings{xie2023two,
  title={Two-stage holistic and contrastive explanation of image classification},
  author={Xie, Weiyan and Li, Xiao-Hui and Lin, Zhi and Poon, Leonard KM and Cao, Caleb Chen and Zhang, Nevin L},
  booktitle={Uncertainty in Artificial Intelligence},
  pages={2335--2345},
  year={2023},
  organization={PMLR}
}

@inproceedings{wang2023counterfactual,
  title={Counterfactual-based Saliency Map: Towards Visual Contrastive Explanations for Neural Networks},
  author={Wang, Han and Zhang, Mingming and Zhang, Song and He, Dongxian and Guo, Yike and Liu, Yang},
  booktitle={2023 IEEE/CVF International Conference on Computer Vision (ICCV)}, 
  year={2023}
}

@article{dhurandhar2018explanations,
  title={Explanations based on the missing: Towards contrastive explanations with pertinent negatives},
  author={Dhurandhar, Amit and Chen, Pin-Yu and Luss, Ronny and Tu, Chun-Chen and Ting, Paishun and Shanmugam, Karthikeyan and Das, Payel},
  journal={Advances in neural information processing systems},
  volume={31},
  year={2018}
}

@inproceedings{kim2018interpretability,
  title={Interpretability beyond feature attribution: Quantitative testing with concept activation vectors (tcav)},
  author={Kim, Been and Wattenberg, Martin and Gilmer, Justin and Cai, Carrie and Wexler, James and Viegas, Fernanda and others},
  booktitle={International conference on machine learning},
  pages={2668--2677},
  year={2018},
  organization={PMLR}
}

@incollection{samek2019towards,
  title={Towards explainable artificial intelligence},
  author={Samek, Wojciech and M{\"u}ller, Klaus-Robert},
  booktitle={Explainable AI: interpreting, explaining and visualizing deep learning},
  pages={5--22},
  year={2019},
  publisher={Springer}
}

@inproceedings{kim2022hive,
  title={HIVE: Evaluating the human interpretability of visual explanations},
  author={Kim, Sunnie SY and Meister, Nicole and Ramaswamy, Vikram V and Fong, Ruth and Russakovsky, Olga},
  booktitle={European Conference on Computer Vision},
  pages={280--298},
  year={2022},
  organization={Springer}
}

@article{ghorbani2020neuron,
  title={Neuron shapley: Discovering the responsible neurons},
  author={Ghorbani, Amirata and Zou, James Y},
  journal={Advances in neural information processing systems},
  volume={33},
  pages={5922--5932},
  year={2020}
}

@InProceedings{zhang25a,
  title = 	 {{Saliency Maps Give a False Sense of Explanability to Image Classifiers}: {A}n Empirical Evaluation across Methods and Metrics},
  author =       {Zhang, Hanwei and Figueroa, Felipe Torres and Hermanns, Holger},
  booktitle = 	 {Proceedings of the 16th Asian Conference on Machine Learning},
  pages = 	 {479--494},
  year = 	 {2025},
  editor = 	 {Nguyen, Vu and Lin, Hsuan-Tien},
  volume = 	 {260},
  series = 	 {Proceedings of Machine Learning Research},
  month = 	 {05--08 Dec},
  publisher =    {PMLR}
}

@article{longo2024explainable,
  title={Explainable Artificial Intelligence (XAI) 2.0: A manifesto of open challenges and interdisciplinary research directions},
  author={Longo, Luca and Brcic, Mario and Cabitza, Federico and Choi, Jaesik and Confalonieri, Roberto and Del Ser, Javier and Guidotti, Riccardo and Hayashi, Yoichi and Herrera, Francisco and Holzinger, Andreas and others},
  journal={Information Fusion},
  volume={106},
  pages={102301},
  year={2024},
  publisher={Elsevier}
}

@article{VILONE202189,
title = {Notions of explainability and evaluation approaches for explainable artificial intelligence},
journal = {Information Fusion},
volume = {76},
pages = {89-106},
year = {2021},
issn = {1566-2535},
doi = {https://doi.org/10.1016/j.inffus.2021.05.009},
author = {Giulia Vilone and Luca Longo},
}

@inproceedings{cntrstv_adapt_nuerips22,
 author = {Wang, Yipei and Wang, Xiaoqian},
 booktitle = {Advances in Neural Information Processing Systems},
 editor = {S. Koyejo and S. Mohamed and A. Agarwal and D. Belgrave and K. Cho and A. Oh},
 pages = {9085--9097},
 publisher = {Curran Associates, Inc.},
 title = {\textquotedblleft Why Not Other Classes?\textquotedblright : Towards Class-Contrastive Back-Propagation Explanations},
 url = {https://proceedings.neurips.cc/paper_files/paper/2022/file/3b7a66b2d1258e892c89f485b8f896e0-Paper-Conference.pdf},
 volume = {35},
 year = {2022}
}

@article{eriksson2025reproducibility,
  title={Reproducibility review of" Why Not Other Classes": Towards Class-Contrastive Back-Propagation Explanations},
  author={Eriksson, Arvid and Israelsson, Anton and Kallhauge, Mattias},
  journal={arXiv preprint arXiv:2501.11096},
  year={2025}
}

@inproceedings{bach2015pixel,
  title={On pixel-wise explanations for non-linear classifier decisions by layer-wise relevance propagation},
  author={Bach, Sebastian and Binder, Alexander and Montavon, Gr{\'e}goire and Klauschen, Frederick and M{\"u}ller, Klaus-Robert and Samek, Wojciech},
  booktitle={PloS one},
  volume={10},
  number={7},
  pages={e0130140},
  year={2015},
  publisher={Public Library of Science}
}

@inproceedings{barkan2023learning,
  title={Learning to explain: A model-agnostic framework for explaining black box models},
  author={Barkan, Oren and Asher, Yuval and Eshel, Amit and Elisha, Yehonatan and Koenigstein, Noam},
  booktitle={2023 IEEE International Conference on Data Mining (ICDM)},
  pages={944--949},
  year={2023},
  organization={IEEE}
}

@inproceedings{barkan2023stochastic,
  title={Stochastic integrated explanations for vision models},
  author={Barkan, Oren and Elisha, Yehonatan and Weill, Jonathan and Asher, Yuval and Eshel, Amit and Koenigstein, Noam},
  booktitle={2023 IEEE International Conference on Data Mining (ICDM)},
  pages={938--943},
  year={2023},
  organization={IEEE}
}

@article{zhang2020explainable,
  title={Explainable recommendation: A survey and new perspectives},
  author={Zhang, Yongfeng and Chen, Xu},
  journal={Foundations and Trends in Information Retrieval},
  volume={14},
  number={1},
  pages={1--101},
  year={2020}
}

@inproceedings{fu2020fairness,
  title={Fairness-aware explainable recommendation over knowledge graphs},
  author={Fu, Zuohui and Xian, Yikun and Gao, Ruoyuan and Zhao, Jieyu and Huang, Qiaoying and Ge, Yingqiang and Xu, Shuyuan and Geng, Shijie and Shah, Chirag and Zhang, Yongfeng and others},
  booktitle={Proceedings of the 43rd international ACM SIGIR conference on research and development in information retrieval},
  pages={69--78},
  year={2020}
}

@article{rudin2019stop,
  title={Stop explaining black box machine learning models for high stakes decisions and use interpretable models instead},
  author={Rudin, Cynthia},
  journal={Nature Machine Intelligence},
  volume={1},
  number={5},
  pages={206--215},
  year={2019},
  publisher={Nature Publishing Group}
}

@article{sokol2020one,
  title={One explanation does not fit all: The promise of interactive explanations for machine learning transparency},
  author={Sokol, Kacper and Flach, Peter},
  journal={KI-K{\"u}nstliche Intelligenz},
  volume={34},
  number={2},
  pages={235--250},
  year={2020},
  publisher={Springer}
}


\clearpage
\appendix
\textbf{\centering\Large{Supplementary Materials: Rethinking Saliency Maps: A Cognitive Human Aligned Taxonomy and Evaluation Framework for Explanations}}\\
\setcounter{section}{0}
\setcounter{secnumdepth}{2}

\section{Appendix Overview}


The appendix provides supplementary materials and detailed analyses that support the findings and discussions presented in the main paper. A summary of its contents is as follows: 

Section~\ref{sec:eval_detailes} outlines additional evaluation details and describes the procedure used to construct the semantic group labels. 

Section~\ref{sec:impl_detailes} explains how existing explanation methods were utilized or adapted to conform to the RFxG framework. 

Section~\ref{sec:add_quan} presents extended quantitative results, including evaluations on the VOC dataset. 

Section~\ref{sec:qual} contains additional qualitative examples illustrating the behavior of explanation methods under the RFxG framework.

Finally, Section~\ref{sec:limitations} discusses the limitations of our work and outlines directions for future research.

\section{Evaluation Details}
\label{sec:eval_detailes}


\paragraph{Dataset Evaluation Details} 
(1) \textbf{IN:} We evaluated on the IN dataset using the original images and their corresponding ground-truth labels. 
(2) \textbf{VOC:} To ensure non-overlapping semantic classes with IN, we manually removed image that belong to a subset of VOC classes (horse, cow, potted plant, and person) that has no overlap with IN classes. For the remaining images in the validation set, we used the top predicted class from the ViT-B model as a ground-truth. 
(3) \textbf{COCO:} The evaluation protocol applied to VOC was also adopted for the COCO dataset.

\paragraph{Semantic Group Construction} 
Semantic groups were constructed based on the WordNet hierarchy. For each class, we first identified whether it belonged to an existing semantic group comprising at least five classes. If such a group was found, the class was assigned to it. Otherwise, we merged the class with neighboring semantic groups until the combined group included at least five classes. If the merged group had a unique superordinate category, we used that as the group label. Otherwise, we concatenated the names of the contributing groups to form the label. This process also used a git repo\footnote{https://github.com/mhiyer/imagenet-hierarchy-from-wordnet/tree/main}.

\section{Method Implementation Details}
\label{sec:impl_detailes}

\paragraph{General Implementation Notes} 
For the evaluation of SHAP, we utilized the GradientSHAP implementation provided by Captum\footnote{\url{https://captum.ai/}}. For Rollout, we employed Gradient-Rollout~\cite{Barkan_2023_ICCV}, an extension that incorporates gradients into the standard attention rollout procedure.

To ensure consistent evaluation across the entire RFxG taxonomy, we adapted a subset of explanation methods, leveraging publicly available implementations where available~\cite{cntrstv_adapt_nuerips22,eriksson2025reproducibility} and applying minimal modifications as needed to align with the specific explanatory objectives of each evaluation setting.
\paragraph{Methods Based on Existing Implementations} 
For CNNs, we used Contrastive GradCAM (GC)~\cite{cntrstv_adapt_nuerips22}. For ViTs, we employed both Contrastive GradCAM-ViT (GCV) and Contrastive Gradient-weighted Attention Rollout~\cite{eriksson2025reproducibility}.

\paragraph{Methods Implemented with Minimal Modifications} 
For explanation methods that lacked native support for contrastive or group-based settings, we developed a general adaptation mechanism that allows alignment with the RFxG taxonomy. Most existing methods operate through a core computation (e.g., backpropagation) applied with respect to a specific class label. We extended this mechanism to support functions over multiple labels, including class differences and group aggregations. Specifically, we applied the core computation over the following functions:
\begin{itemize}
\item \textbf{Class-Contrastive Methods:} The explanation is computed with respect to the score difference between class $A$ and class $B$, i.e., $f_A - f_B$. 
\item \textbf{Class-to-Group Contrastive Methods:} The explanation is computed using the difference between the score for class $A$ and the sum of scores for all classes in a contrastive group $G_B$, i.e., $f_A - \sum_{c \in G_B} f_c$. 
\item \textbf{Pointwise Group Methods:} The explanation is computed with respect to the aggregated score of all classes within a semantic group $G$, i.e., $\sum_{c \in G} f_c$. 
\item \textbf{Contrastive Group Methods:} The explanation is computed based on the difference between the sum of scores for classes in group $G_A$ and those in group $G_B$, i.e., $\sum_{{c_1} \in G_A} f_{c_1} - \sum_{{c_2} \in G_B} f_{c_2}$. 
\end{itemize}

This adaptation strategy enables consistent application of explanation techniques under varying reference frames and semantic granularities as required by the RFxG framework.

\paragraph{Example: Customizing Contrastive-GradCAM.} \textbf{Note:} The implementation presented here is not based on the official method from~\cite{cntrstv_adapt_nuerips22}, but rather serves as an illustrative example of how we adapted GC to align with the RFxG framework.
This implementation of Contrastive-GradCAM generalizes the GC method~\cite{selvaraju2017grad} to generate explanations that highlight features distinguishing a target class $A$ from an alternative class $B$. Instead of computing gradients with respect to the class score $f_A$ alone, this version of GC computes gradients with respect to the contrastive score:
\begin{equation}
    s_{A,B} = f_A - f_B,
\end{equation}
where $f_A$ and $f_B$ are the pre-softmax logits for classes $A$ and $B$, respectively.

Let $A^k$ denote the $k$-th activation map from the final convolutional layer. The corresponding importance weight $\alpha_k^{A,B}$ is computed via global average pooling over the gradients:
\begin{equation}
    \alpha_k^{A,B} = \frac{1}{Z} \sum_{i,j} \frac{\partial s_{A,B}}{\partial A^k_{i,j}},
\end{equation}
where $Z$ is the number of spatial positions and $(i,j)$ indexes spatial coordinates. Unlike standard GC, we omit the final ReLU operation to preserve negative contributions, as these are considered informative in the contrastive setting~\cite{cntrstv_adapt_nuerips22}. The resulting contrastive saliency map is given by:
\begin{equation}
    M_{\text{con}}^{A,B} = \sum_k \alpha_k^{A,B} A^k.
\end{equation}
This formulation enables the explanation to focus on regions that are specifically important for differentiating class $A$ from class $B$.

\section{Additional Quantitative Results}
\label{sec:add_quan}
Table~\ref{tab:only_voc_results} presents further results for the VOC datasets. These results follow similar trends to Tab.~\ref{tab:results}.
\begin{table}[t!]
\centering
\caption{Evaluation results across datasets, models, and explanation methods using our metrics: CCS, CGC, PGS, and CGS. Higher scores indicate better performance. For every metric in the table, the results were provided using a method suitable for the metric's purpose. For example: for evaluating GC on CCS, we used Contrastive-GradCAM~\cite{cntrstv_adapt_nuerips22}.}
\label{tab:only_voc_results}
\scalebox{0.8}{
\begin{tabular}{lll|cccc}
\toprule
Dataset & Model & Method & CCS $\uparrow$ & CGC $\uparrow$ & PGS $\uparrow$ & CGS $\uparrow$ \\
\midrule
\multirow{20}{*}{VOC}
& \multirow{5}{*}{CN} & IIA & \textbf{25.17} & \textbf{5.19} & \textbf{46.39} & \textbf{35.61} \\
& & GC & \underline{21.96} & \underline{4.79} & \underline{42.43} & \underline{32.64} \\
& & SC & 21.52 & 4.52 & 41.74 & 31.87 \\
& & SHAP & 16.20 & 3.26 & 32.19 & 26.86 \\
& & IG & 13.56 & 2.93 & 27.69 & 23.20 \\
\cmidrule(lr){2-7}
& \multirow{5}{*}{RN} & IIA & \textbf{25.45} & \textbf{5.23} & \textbf{46.48} & \textbf{36.35} \\
& & GC & \underline{22.79} & \underline{4.65} & \underline{42.14} & \underline{32.97} \\
& & SC & 22.16 & 4.39 & 40.12 & 32.03 \\
& & SHAP & 16.38 & 3.41 & 32.27 & 27.00 \\
& & IG & 13.76 & 2.79 & 27.59 & 23.22 \\
\cmidrule(lr){2-7}
& \multirow{5}{*}{ViT-B} & IIA & \textbf{19.64} & \textbf{4.57} & \textbf{39.60} & \textbf{29.82} \\
& & TAttr & \underline{18.47} & \underline{4.24} & \underline{36.47} & \underline{27.70} \\
& & GAE & 18.24 & 4.01 & 35.84 & 27.38 \\
& & Rollout & 14.79 & 3.02 & 29.20 & 22.36 \\
& & GCV & 12.93 & 2.81 & 26.66 & 20.41 \\
\cmidrule(lr){2-7}
& \multirow{5}{*}{ViT-S} & IIA & \textbf{19.63} & \textbf{4.59} & \textbf{39.36} & \textbf{29.64} \\
& & TAttr & \underline{18.31} & \underline{4.22} & \underline{35.56} & \underline{27.01} \\
& & GAE & 18.03 & 4.06 & 35.00 & 26.61 \\
& & Rollout & 14.88 & 3.08 & 29.15 & 22.36 \\
& & GCV & 12.56 & 2.73 & 26.39 & 20.39 \\
\bottomrule
\end{tabular}}
\end{table}

\section{Qualitative Results}
\label{sec:qual}
\begin{figure}[t]
  \centering
  \includegraphics[width=0.5\textwidth]{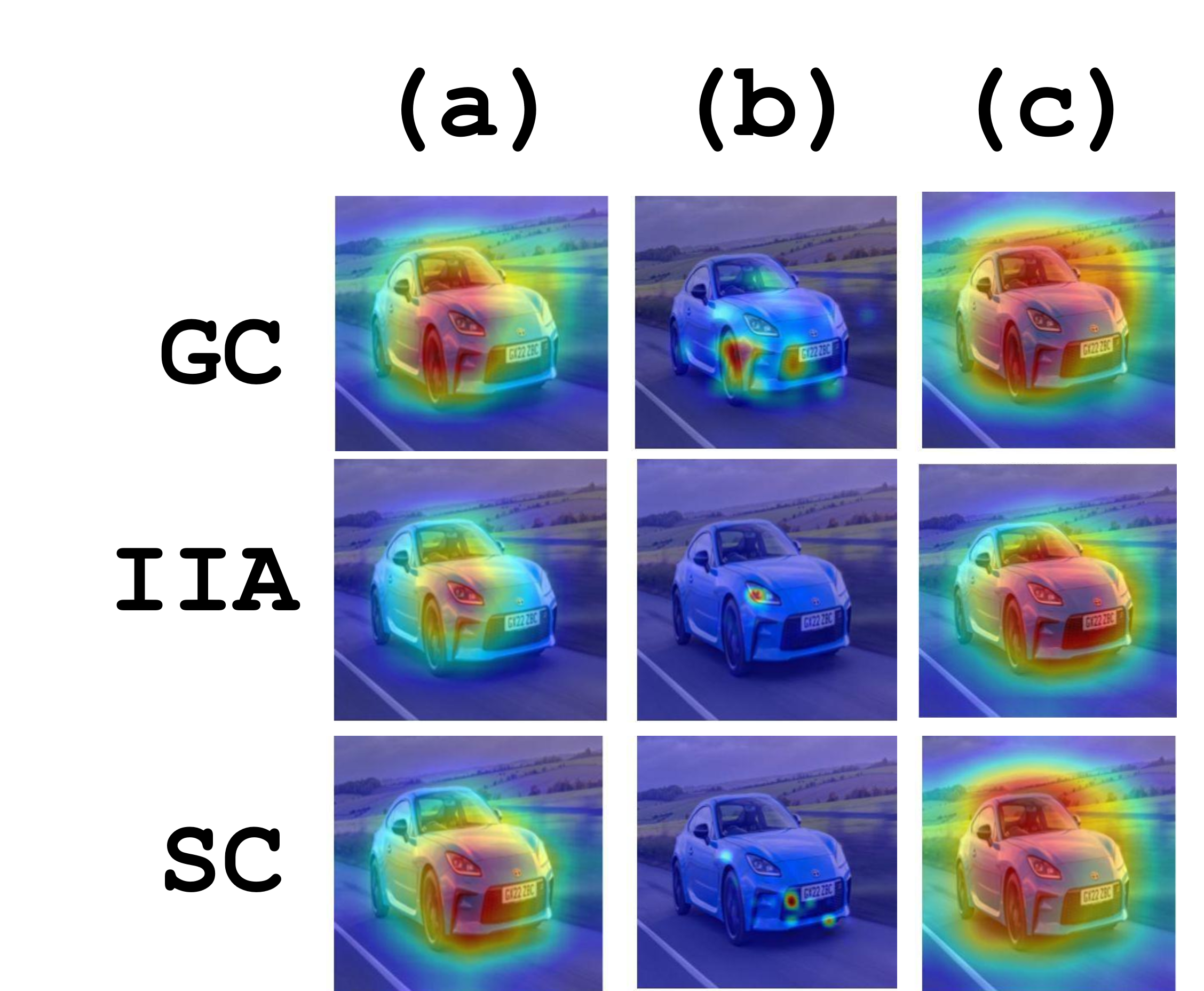}
  \caption{Saliency maps for a sport car using different types of questions that reflect our taxonomy with different methods: (a) Class Contrastive. (b) Class-Group Contrastive. (c) Pointwise Group.}
  \label{fig:quali}
\end{figure}

Figure~\ref{fig:quali} presents a qualitative comparison of the explanation maps obtained by the methods GC,IIA, and SC. These maps were created using the following setup: A is 'sports car, sport car', B is 'cab, hack, taxi, taxicab' and the group named Car with the following classes: 'sports car','racer','model T', 'minivan', 'limousine', 'jeep', 'convertible', 'cab', 'beach wagon', and 'ambulance'.

\section{Limitations and Future Work}
\label{sec:limitations}




While our RFxG framework provides a significant advancement in user-aligned saliency map evaluation, several limitations warrant discussion alongside promising future directions. These considerations do not undermine our core contributions but rather highlight opportunities to extend the impact of our principled approach to explanation evaluation.

\subsection{Methodological Considerations}

Perturbation-based evaluation framework, while widely adopted in the XAI literature \cite{petsiuk2018rise,samek2017evaluating}, can inherit certain methodological constraints. 
For instance, our black pixel masking approach we employed could theoretically introduce distribution shift artifacts when removing salient regions. To address this concern, we conducted extensive ablation studies with alternative masking techniques, including Gaussian blur and texture-preserving inpainting. While these alternatives mitigate distribution shift concerns, they yielded qualitatively and quantitatively similar results across all RFxG metrics, confirming that our framework's insights are not artifacts of the specific perturbation methodology. Future work could explore more sophisticated perturbation approaches that better preserve local image statistics while maintaining computational efficiency.

\subsection{Scope and Generalizability}

Our current implementation focuses on image classification tasks, leaving open opportunities for extension to other domains. The RFxG framework naturally extends to vision-based recommendation systems, where explaining why item A was recommended over item B (contrastive reference frame) at varying levels of abstraction (granularity) could enhance user trust and satisfaction. For instance, in fashion recommendation, our framework could elucidate whether a recommendation stems from fine-grained attributes (specific pattern details) or broader category features (general garment type).

In natural language processing, the RFxG taxonomy offers promising avenues for advancing text explanation methods. Current saliency approaches for text often lack explicit reference frames, making it difficult to distinguish between explanations answering "Why this sentiment?" versus "Why positive and not negative?". Applying our framework to transformer-based language models could yield more precise explanations aligned with user queries, particularly valuable in applications like legal document analysis or medical report interpretation where contrastive reasoning is essential.

\subsection{Integration with Concept-Based Explanations}

While our work establishes a foundation for user-aligned evaluation of pixel-space saliency maps, an exciting direction involves bridging the gap between low-level visual explanations and high-level conceptual reasoning. The RFxG framework could be extended to evaluate concept-based explanation methods like TCAV \cite{kim2018interpretability} by formalizing how concepts operate across reference frames (e.g., "Why concept X supports class A versus class B?") and granularities (e.g., "How does concept X operate at different semantic levels?"). This integration would create a continuum from pixel-level to concept-level explanations, addressing a critical gap in current XAI research.

Furthermore, our metrics could be adapted to evaluate explanations in multimodal settings, where understanding the interplay between visual and textual explanations becomes crucial. For instance, in vision-language models, determining whether a model relies on appropriate visual evidence versus language priors when answering contrastive questions represents an important application of our framework.

\subsection{Broader Impact and Deployment Considerations}

As our framework moves toward real-world deployment, particularly in high-stakes domains like medical imaging and autonomous systems, additional considerations emerge. In medical applications, for example, the appropriate granularity for explanations may differ between clinicians (who might prefer organ-level group explanations) and patients (who might benefit from more fine-grained lesion-specific explanations). Future work should investigate domain-specific adaptations of RFxG that respect the unique requirements and constraints of different application areas.

Moreover, while our current metrics focus on faithfulness to model behavior, complementary metrics assessing explanation utility for specific tasks (e.g., model debugging versus user education) represent a valuable extension. Developing a comprehensive evaluation suite that balances faithfulness with task-specific utility would further strengthen the practical impact of explanation methods.

In conclusion, rather than representing weaknesses of our framework, these limitations highlight fertile ground for extending the RFxG taxonomy's impact across AI subfields and applications. By addressing these directions, the research community can build upon our foundation to create explanation systems that are not only technically sound but also genuinely useful for diverse users across myriad contexts. Our work provides the conceptual scaffolding necessary to transform saliency maps from ambiguous visualizations into precise, user-aligned communication tools, a critical step toward trustworthy and interpretable AI systems.



\end{document}